%% file: neurips_2022.tex
\documentclass{article}

    \usepackage[preprint]{neurips_2022}

\usepackage[utf8]{inputenc} % allow utf-8 input
\usepackage[T1]{fontenc}    % use 8-bit T1 fonts
\usepackage{hyperref}       % hyperlinks
\usepackage{url}            % simple URL typesetting
\usepackage{booktabs}       % professional-quality tables
\usepackage{amsfonts}       % blackboard math symbols
\usepackage{nicefrac}       % compact symbols for 1/2, etc.
\usepackage{microtype}      % microtypography
\usepackage{xcolor}         % colors

\usepackage{bm}

\title{Generalized Uncertainty of Deep Neural Networks: Taxonomy and Applications}

\author{%
  Chengyu Dong \\
  University of California, San Diego\\
  \texttt{cdong@eng.ucsd.edu} \\
}

\newcommand{\smallsection}[1]{\textbf{#1.~~~~}}

\begin{document}

\maketitle

\input{sections/0-abstract.tex}
\input{sections/1-intro.tex}
\input{sections/2-preliminary.tex}

\input{sections/3-uncertainty.tex}
\input{sections/4-usage.tex}
% \input{neurips2022/sections/5-study.tex}
\input{sections/6-conclusion.tex}

%%%%%%%%%%%%%%%%%%%%%%%%%%%%%%%%%%%%%%%%%%%%%%%%%%%%%%%%%%%%
% \newpage
\bibliography{neurips_2022}
\bibliographystyle{neurips_2022}

%%%%%%%%%%%%%%%%%%%%%%%%%%%%%%%%%%%%%%%%%%%%%%%%%%%%%%%%%%%%

% \newpage
% \appendix

% \section{Appendix}
% Optionally include extra information (complete proofs, additional experiments and plots) in the appendix.
% This section will often be part of the supplemental material.

\end{document}

%% file: sections/0-abstract.tex
\begin{abstract}
Deep neural networks have seen enormous success in various real-world applications.
% , frequently surpassing human performance in specific tasks. 
Beyond their predictions as point estimates, increasing attention has been focused on quantifying the uncertainty of their predictions. In this review, we show that the uncertainty of deep neural networks is not only important in a sense of interpretability and transparency, but also crucial in further advancing their performance, particularly in learning systems seeking robustness and efficiency. We will generalize the definition of the uncertainty of deep neural networks to any number or vector that is associated with an input or an input-label pair, and catalog existing methods on ``mining'' such uncertainty from a deep model. We will include those methods from the classic field of uncertainty quantification as well as those methods that are specific to deep neural networks. We then show a wide spectrum of applications of such generalized uncertainty in realistic learning tasks including robust learning such as noisy learning, adversarially robust learning; data-efficient learning such as semi-supervised and weakly-supervised learning; and model-efficient learning such as model compression and knowledge distillation.
\end{abstract}

%% file: sections/1-intro.tex
\section{Introduction}
Despite the vast success of deep neural networks, their decision process is hard to interpret and is known as a black box. In real-world applications, it is necessary that a decision system is not only accurate but also trustworthy, in the sense that it must know when it is likely to make errors~\citep{Guo2017OnCO}. The interpretability and transparency of the decision process of deep neural networks have thus gained increasing attention, around which the pivot is often a reliable uncertainty measure that users can judge and manage the decisions. A variety of uncertainty measures and strategies to improve them have been developed so far.

In this review, we show that the uncertainty estimates of deep neural networks are important not only in improving their trustworthiness, but also in further advancing their performance, particularly in terms of robustness and efficiency. We will first review the possible uncertainty estimates we can leverage for deep neural networks. These include the classic definition of uncertainty, for example, the maximum probability or entropy of the predictive distribution, as well as strategies to improve it such as a diverse form of ensemble techniques. These ensemble techniques are mostly unique to deep neural networks, which either utilize the specific design in the network architecture such as MC-dropout~\citep{Gal2016DropoutAA}, or utilize the distinct optimization process of deep neural networks such as Snapshot ensemble~\citep{Huang2017SnapshotET}. 

We will further investigate the uncertainty estimates that are defined beyond the predictive distribution. For this, we generalize the definition of uncertainty from an estimate associated with a model prediction to any number or vector associated with a data example, where the label can either be provided or missing. We see that under such a definition, multiple intriguing properties of deep neural networks can be leveraged to define an uncertainty estimate. These include measures based on inference dynamics of deep neural networks such as prediction depth~\citep{Baldock2021DeepLT}, and measures based on training dynamics of deep neural networks such as learning order~\citep{Arpit2017ACL, Hacohen2020LetsAT}.

Finally, we discuss the potential applications of these uncertainty estimates in various learning problems, particularly when robustness and efficiency are of major interest. We show that in realistic datasets where the labels are expensive to obtain or the label noise is pervasive, reliable uncertainty estimates can be utilized to improve the performance greatly. We also show that uncertainty estimates can be utilized to enhance the performance of deep neural networks under adversarial attacks~\citep{Goodfellow2015ExplainingAH}. 
We then move to learning problems where the computation cost is prohibitive. We show that reliable uncertainty estimates can improve both the training and inference efficiency of deep neural networks, when utilized in advanced efficient learning techniques such as knowledge distillation~\citep{Hinton2015DistillingTK} and adaptive inference time~\citep{Graves2016AdaptiveCT}.

% Deep double descent~\citep{Nakkiran2020DeepDD}

% Not only an interpretability concern, but also productivity. May be particular interest to those working on applied problems.

This review will be structured as follows. In Section~\ref{sect:preliminary}, we will introduce the necessary background such as the design, training and inference of deep neural networks, as well as the generalized definition of uncertainty. In Section~\ref{sect:uncertainty}, we will review various existing uncertainty estimates under such a definition, along with their problems and the strategies to improve them. In Section~\ref{sect:usage}, we demonstrate how we can use the uncertainty estimates of deep neural networks to advance their performance in robust and efficient learning. Finally, Section~\ref{sect:conclusion} concludes our review and illuminates the potential opportunities in this direction.

%% file: sections/2-preliminary.tex
\section{Preliminaries}
\label{sect:preliminary}
\subsection{Deep neural networks}
We consider a supervised learning setting where a deep neural network $f_\theta$ is defined as a function mapping from input domain $\mathcal{X}$ to output domain $\mathcal{Y}$, namely $f_\theta: \mathcal{X} \to \mathcal{Y}$. A deep neural network usually consists of multiple feed-forward layers, where each layer $l$ is parameterized by a set of weights $\theta_l$. We denote the function mapping of the deep neural network up to layer $L$ as $f_{\theta_{1:L}}$. With a slight abuse of notation, we denote the function mapping of all layers, or the entire deep neural network as $f_\theta$. Note that it is not necessarily the case that $f_{\theta_{1:L}}(x) \in \mathcal{Y}$, namely the output of an intermediate layer can have higher or lower dimensionality than the output domain. During inference, given any unseen example input-label pair $(x,y)$, where $x\in \mathcal{X}$ and $y\in \mathcal{Y}$, a deep neural network accepts $x$ and produces a prediction $f_\theta(x)$, where we expect $f_\theta(x) = y$.

$$
y = f_\theta(x)
$$

Training a deep neural network usually requires a training sample, namely a set of input-label pairs $\mathcal{D} = \{(x_i, y_i)\}_{i=1}^N$. The most commonly used method to train a deep neural network is empirical risk minimization. We first define a distance function in the output domain $l: \mathcal{Y}\times \mathcal{Y} \to \mathbb{R}$, which we typically refer as the loss function. It is desired that such a distance function suffices $l(y_1, y_2)\ge 0$ for any $y_1, y_2 \in \mathcal{Y}$ and $l(y, y) = 0$ for any $y \in \mathcal{Y}$. Empirical risk minimization can be defined as
\begin{equation}
    \theta^* = \arg\min_\theta \sum_{(x,y) \in \mathcal{D}} l(f_\theta(x), y),
\end{equation}
namely we seek an optimal set of model weights that minimize the distance between the network outputs and the labels for all training examples.

To solve such a minimization problem, the typically used optimization method is gradient descent with multiple updates. For a total of $M$ updates, we repeatedly calculate the gradient of each weight with respect to the minimization objective and update the weights in the opposite direction of the gradient. In its simplest form, the update rule can be expressed as
\begin{equation}
    \theta_{t+1} = \theta_t - \alpha \nabla_\theta \sum_{(x,y) \in \mathcal{D}} l(f_{\theta_t}(x), y),
\end{equation}
where $\alpha$ is a scalar and is typically referred to as the learning rate. Here we have denoted the network weights after $t$ updates as $\theta_t$.

\subsection{Definition of the generalized uncertainty}
\label{sect:definition}
We define the uncertainty of a deep neural network as a function that maps a data example to any real number or vector, namely $c_\theta: \mathcal{X}\times \mathcal{Y} \to \mathbb{R}^K$, where $c_\theta$ means that such a function mapping is defined by the network $f_\theta$. Note that sometimes such a function mapping can take the input only, with the label missing, in which case we denote it as $c_\theta: \mathcal{X} \to \mathbb{R}^K$, with a slight abuse of notation.

We note that such a definition can cover multiple classical definitions of uncertainty. For example, in a multi-class classification setting where each input is associated with a label selected from a set of $K$ classes, a deep neural network's confidence of its prediction is simply defined as the maximum probability mass in its output vector, namely $c_\theta(x) = \max_k f_\theta(x)[k]$, where $f_\theta(x)[k]$ denotes the $k$-th entry of the prediction vector $f_\theta(x)$. Here we have in fact interpreted the prediction vector in a probabilistic sense. Let $\mathbf{1}_{k=k'} \in \mathcal{Y} \subseteq \mathbb{R}^K$ denotes the vector where only the $k'$-th entry is $1$ and others are $0$, we then have $\Omega = \{\mathbf{1}_{k=1}, \mathbf{1}_{k=2}, \cdots, \mathbf{1}_{k=K}\}$ defined as the sample space, and $f_\theta(x)[k']$ denotes the probability measure of the singleton set $\{\mathbf{1}_{k=k'}\}$. One may recognize that this is the standard definition of a categorical distribution. 

Note that in order to use the prediction vector as a valid probability measure, it is desired that the prediction vector lies in a $(K-1)$-dimensional simplex, namely $f_\theta(x) \in \Delta_{K-1} := \{y\in [0,1]^K ~|~ \|y\|_1 =1 \}$. We will review more typical definitions of the uncertainty of deep neural networks in later sections.

% A general definition: $P_\theta(y | x)$, where $y$ is the prediction and $x$ is the input.

% classic definition of Uncertainty and usage
% E.g. Confidence (mention in the later that for deep neural networks, uncertainty might need to be calibrated) and OOD detection

%% file: sections/3-uncertainty.tex
\section{Uncertainty Estimation of Deep Neural Networks}
\label{sect:uncertainty}
Standard deep neural networks are usually deterministic inference systems, namely the prediction on a given input will not vary for multiple forward passes. It is possible to inject inherent randomness into a deep model's inference process by specifying a prior distribution over its weights $\theta$. By the Bayesian theorem, the uncertainty of a prediction can then be captured by the posterior distribution of the prediction conditioned on the model weights. Such a deep neural network variation is called a Bayesian Neural Network (BNN)~\citep{Denker1987LargeAL, Tishby1989ConsistentIO, Buntine1991BayesianB}. It is rather obvious that the inference of such neural networks would be computationally expensive, as multiple samplings of the model weights are required for one single inference. There have been a variety of attempts to make BNNs computationally tractable, such as Laplace approximation~\citep{Bridle2011BayesianMF} and Markov chain Monte Carlo (MCMC) methods~\citep{Neal1995BayesianLF}.

In this review, we will focus only on the uncertainty of standard deterministic deep neural networks, while skipping recent works on BNNs. The reasons are two folds. First, in practice, BNNs are still difficult to implement and computationally slow~\citep{Lakshminarayanan2017SimpleAS}. Second and more importantly, in this review, we are not seeking the role of uncertainty in the transparency of the model decision process, but rather in aiding the model training. For example, a typical application of uncertainty is to select high-quality unlabeled data that is most relevant to the current task from a large corpus. We will discuss more of such applications in the following sections.

% \note{Only data uncertainy (because most cases used to select data), not interested in model uncertainy (e.g. BNNs)}

\subsection{Classic uncertainty}
In Section~\ref{sect:definition}, we have mentioned the classic way to interpret the predictive vector of a deep neural network as a probability distribution. In practice, to create a valid probabilistic vector, we often need to apply a non-linear activation function on top of the network outputs. Let the original output vector of a network as $z\in \mathbb{R}^K$, and $\sigma: \mathbb{R}^K \to [0,1]^K$ as a non-linear function. We will then make $f_\theta(x) = \sigma(z)$ for an input $x$, where $z$ is often referred as the logits. For example, in a binary classification task, we will use the sigmoid function, namely 
$$
\sigma(z) = \frac{1}{1 + \exp(-z)},
$$
and in a multi-class classification task, we will use the softmax function, namely 
$$
\sigma(z)[k] = \frac{\exp(z[k])}{\sum_{k=1}^K \exp(z[k])}.
$$

A valid probabilistic vector can be directly interpreted as the uncertainty of the model prediction. One can also transform such an uncertainty vector into a number, which will be particularly handy in obtaining the ranking of a set of data examples based on an uncertainty measure. Various transformations are used in the literature, typically including maximum probability, entropy, and margin. 
\begin{itemize}
    \item Maximum probability: $c_\theta(x) = \max_k f_\theta(x)[k]$,
    \item Entropy: $c_\theta(x) = - f_\theta(x) \cdot \log f_\theta(x) $,
    \item Margin: $c_\theta(x) = \max_k f_\theta(x)[k] - \max_{k\ne k'} f_\theta(x)[k]$, where $k' = \arg\max_k f_\theta(x)[k]$.
\end{itemize}

\subsection{Evaluation of classic uncertainty}
The desired properties of an uncertainty estimate may vary significantly across the tasks of interest, sometimes even orthogonally. Consequently, a variety of uncertainty evaluation metrics exists in the literature. Here we roughly group them into three types based on the property desired.

\smallsection{Recovering the true distribution}
Assuming the data examples are sampled from a joint distribution $P_{X,Y}(x,y) = P_X(x) P_{Y|X}(y|x)$ defined over $\mathcal{X} \times \mathcal{Y}$. Given any input $x$, it is desired that the predictive uncertainty, namely the probabilistic vector $f_\theta(x)$ produced by a deep neural network, can recover the true conditional distribution $P_{Y|X}(y|x)$. Note that such recovery is a stricter concern compared to accuracy~\citep{Lakshminarayanan2017SimpleAS}. A network's prediction can be very accurate, yet significantly deviate from the true conditional distribution. A typical example is that the network always outputs the one-hot label on an input $x$.

To evaluate the quality of the recovery, one can use commonly seen loss functions such as negative log-likelihood (NLL) loss and mean squared error (MSE). Interestingly, it is known in meteorology that these two loss functions have good properties~\footnote{MSE is also known as the Brier score in meteorology}, or they are known as proper scoring rules~\citep{Gneiting2007StrictlyPS}. A proper scoring rule is one loss function where $l(p, q) \ge l(q, q)$ with equality if and only if $p = q$. This means that minimizing the distance between the predictive uncertainty and one-hot labels can in fact recover the true conditional distribution. To see that for NLL, we can apply the Gibbs inequality, namely
\begin{equation}
    -\mathbb{E}_{(x,y)\sim P(x,y)} \mathbf{1}_{y} \cdot \log f_\theta(x) = -\mathbb{E}_{x\sim P(x)} P(y|x)\cdot \log f_\theta(x) \ge  -\mathbb{E}_{x\sim P(x)} P(y|x)\cdot \log P(y|x),
\end{equation}
where the equality establishes if and only if $f_\theta(x) = P(y|x)$. To see that for MSE, we can decompose the MSE and find that
\begin{equation}
    \|\mathbf{1}_y - f_\theta(x)\|_2^2 = \|\mathbf{1}_y - P(y|x)\|_2^2 + \|P(y|x) - f_\theta(x)\|_2^2 \ge \|\mathbf{1}_y - P(y|x)\|_2^2,
\end{equation}
where the equality establishes if and only if $f_\theta(x) = P(y|x)$. 
Therefore, in practice when there are only one-hot labels available, we can still use NLL or MSE to quantify the quality of the uncertainty in terms of its recovery to the true conditional distribution.

\smallsection{Calibration}
Loosely speaking, calibration of deep neural networks implicates the network uncertainty should reflect the probability that it makes errors~\citep{Guo2017OnCO}. Here we are only interested in the network's argmax prediction, which we denote as $y_\theta = \arg\max_k f_\theta(x)[k]$ for simplicity. We are also only interested in the uncertainty as a real number between $0$ and $1$, which reflects the probability of the pointwise prediction $y_\theta$ being correct. Usually, this is simply the maximum probability, namely $c_\theta(x) = \max_k f_\theta(x)[k]$. Formally, calibration then desires that~\citep{Guo2017OnCO}
\begin{equation}
    P(y_\theta = y ~|~ c_\theta(x) = p) = p,~\forall~p\in[0, 1].
\end{equation}
Here the left denotes the accuracy of the network prediction on all examples where the network reports uncertainty of $p$, and the right is the value of the uncertainty. Therefore, calibration means that the uncertainty of a prediction should genuinely match the probability of correctness of such prediction.
It may help the understanding of calibration if we interpret the probability $P$ as the limit of the frequency. Therefore, an alternative way to formalize calibration is~\citep{Kuleshov2018AccurateUF},
\begin{equation}
    \lim_{N\to \infty}  \frac{\sum_{i=1}^N1(y_\theta = y_i)\cdot 1(c_\theta(x_i) = p)}{\sum_{i=1}^N1(c_\theta(x_i) = p)} = p,
\end{equation}
where $1(\cdot)$ is the indicator function.

A legacy issue is that, calibration of deep neural networks is in fact a weaker requirement of the model uncertainty compared to the definition of calibration in standard statistical terminology~\citep{Zadrozny2001ObtainingCP}, which can be denoted as
\begin{equation}
    P(y \in \cdot | f_\theta(x) = \bm{p}) = \bm{p},~\forall~\bm{p}\in \Delta_{K-1},
\end{equation}
namely for any predictive distribution of the network, the label should distribute exactly as that predictive distribution. Note that, although a true model producing the true conditional distribution is certainly calibrated, a (statistically) calibrated model does not necessarily have to  recover the true conditional distribution and a model that is not close to the true model can nevertheless be calibrated~\citep{Vaicenavicius2019EvaluatingMC}.

To measure the uncertainty quality in terms of calibration, one can use the difference in expectation between the accuracy and the uncertainty, namely
\begin{equation}
    \label{eq:ece}
    \mathbb{E}_{p}\left[ \left| P(y_\theta=y ~| ~c_\theta(x) = p) - p \right | \right].
\end{equation}
In practice, one can report the Expected Calibration Error (ECE)~\citep{Naeini2015ObtainingWC}, which approximates the expectation by binning. Specifically, the predictions on all test examples are partitions into several equally-spaced bins, where predictions with similar uncertainty will be assigned into the same bin. We can then approximate $P(y_\theta=y ~| ~c_\theta(x) = p)$ in Equation~(\ref{eq:ece}) by the faction of correct predictions in each bin, and approximate $p$ in Equation~(\ref{eq:ece}) by the average uncertainty in this bin.

When the calibration of all-class predictions instead of the argmax prediction is of interest, one can also use the Static Calibration Error (SCE)~\citep{Nixon2019MeasuringCI} to quantify the uncertainty quality. All-class calibration measures are often more effective in assessing the calibration error~\citep{Nixon2019MeasuringCI}. Other variation includes the adaptive Expected Calibration Error (aECE)~\citep{Nixon2019MeasuringCI}, which partitions the predictions into several bins with an equal number of predictions in each bin. This can be more robust to the number of bins~\citep{Patel2021OnmanifoldAD}, since the uncertainty distribution is often far from a uniform one~\citep{Guo2017OnCO}, and the number of bins is critical in evaluating the calibration genuinely~\citep{Kumar2019VerifiedUC}.

% Calibration error, multi-class calibration, perfect calibration (same as NLL)
% * Partially recover the true distribution
% * Relation between calibration error and total calibration (`evaluate..`)
%     * Total calibration -> calibration error = 0
%     * calibration error = 0 !-> Total calibration
%     * Is there a decomposition?
    
\smallsection{Ordinal ranking}
In many practical settings, the absolute distance between model uncertainty and the probability of correctness is often not necessary. Instead, the ranking of a set of predictions based on the uncertainty measure is more important. The major objective here is to distinguish correct from incorrect predictions. Therefore it is desired that correct predictions have higher confidence estimates than incorrect predictions. Formally, a prefect ordinal ranking means that~\citep{Moon2020ConfidenceAwareLF}
\begin{equation}
    c_\theta(x_i) \le c_\theta(x_j) \Longleftrightarrow P(y_{\theta, i} = y_i | x_i) \le P(y_{\theta, j} = y_j | x_j).
\end{equation}

To measure the uncertainty quality in terms of ordinal ranking, one can specify an uncertainty threshold, such that predictions with uncertainty above this threshold are regarded as correct predictions. However, this often rises the problem of a trade-off between false negatives and false positives~\citep{Hendrycks2017ABF}. For a threshold-free evaluation, one can use metrics such as the Area Under the Receiver Operating Characteristic curve (AUROC)~\citep{Davis2006TheRB}, the Area Under the Precision-Recall curve (AURC)~\citep{Manning2002FoundationsOS} and the Area under the Risk-Coverage curve (AURC)~\citep{Geifman2019BiasReducedUE}. The underlying idea of all these measures is to aggregate the accuracy under all possible thresholding of the set of predictions.

Ordinal ranking has a wide variety of applications in practice. For example, uncertainty with accurate ranking can effectively identify those examples that come from a distribution substantially different from the distribution that the model is trained on. Such examples are known as the out-of-distribution (OOD) examples. Ordinal ranking is also important in active learning~\citep{Settles2009ActiveLL}, where the goal is to build a model knowing which examples should be labeled to improve its performance. Quality uncertainty ranking can thus greatly reduce human labeling efforts. Ordinal ranking is also crucial in selective classification~\citep{Geifman2017SelectiveCF} or failure prediction~\citep{Hendrycks2017ABF, Hecker2018FailurePF}, where the goal is to reject some predictions in test time that are likely to be incorrect. The rejected examples can be passed on to backup inference systems or humans, such that the overall prediction accuracy can be greatly improved. 
We will discuss more applications in later sections and frequently revisit the ranking measure of uncertainty.

\subsection{Problems and improvement of classic uncertainty}
It is well-known that the classic uncertainty of deep neural networks may have major drawbacks. It may be poorly calibrated~\citep{Guo2017OnCO}, yields inconsistent ranking~\citep{Corbire2019AddressingFP}, and is vulnerable to perturbations such as adversarial attack~\citep{Szegedy2014IntriguingPO, Goodfellow2015ExplainingAH} and dataset shifts~\citep{Hendrycks2017ABF, Ovadia2019CanYT}.

To overcome these drawbacks of classic uncertainty, a variety of methods have been proposed, which can be roughly divided into two veins. The first vein is called post-processing, namely the uncertainty is rectified after a model is trained. In this case, a validation set is often needed. The second vein can be broadly referred to as regularization, namely the training process is modified to take into consideration. not only the accuracy but also the uncertainty.

\smallsection{Post-processing} The very first and also the simplest post-processing method for deep neural networks is probably temperature scaling~\citep{Guo2017OnCO}. The idea is to insert a hyperparameter called temperature in the softmax function of a trained model and fine-tune it on a validation set. The softmax function now becomes
\begin{equation}
    \sigma(z; T)[k] = \frac{\exp(z[k]/T)}{\sum_{k=1}^K \exp(z[k]/T)}.
\end{equation}
When $T$ is larger, the new predictive distribution will become softer (higher entropy), which may alleviate the typical problem of deep neural network's uncertainty, namely being over-confident and inappropriately close to the one-hot label. Temperature scaling has been shown to be quite effective in improving the uncertainty, particularly in terms of calibration, outperforming more sophisticated calibration methods such as histogram binning~\citep{Zadrozny2001ObtainingCP}, Isotonic regression~\citep{Zadrozny2002TransformingCS} and Platt scaling~\citep{Platt1999ProbabilisticOF}. On top of its simplicity and effectiveness, another reason for the popularity of temperature scaling is probably that temperature scaling will preserve the model's accuracy albeit improving the uncertainty, because scaling the logits by a scalar will not change the argmax of the predictive vector. Note that despite its success in calibration, temperature scaling cannot improve the ordinal ranking of the model uncertainty. The reason is similar as the logit scaling is universal to all examples.

Many variations of temperature scaling have been proposed to further enhance its effectiveness. For example, one can perform the temperature scaling for each class in a one-vs-all manner, despite sacrificing the accuracy since the argmax label is no longer preserved~\citep{Kull2019BeyondTS}. Multiple temperature scaling methods can be ensembled to achieve better uncertainty quality~\citep{Zhang2020MixnMatchEA}. One can also combine temperature scaling with other calibration methods such as histogram binning to get a theoretical guarantee on the calibration error~\citep{Kumar2019VerifiedUC}. \citet{Liang2018EnhancingTR} observed that adding small adversarial perturbation to the input after temperature scaling can further improve the ranking of the uncertainty and thus better separate in- and out-of-distribution examples. When there are data from multiple domains available, one can learn and predict the most proper temperature when encountering an unseen example that is likely shifted from existing distributions~\citep{Yu2022RobustCW}.

\smallsection{Regularization} The standard deep neural network training protocol may already have some specific design that favors the learning of uncertainty. For example, as also mentioned before, the typically used loss functions such as NLL and MSE are in fact proper scoring rules, which can recover the true conditional distribution when minimized~\citep{Lakshminarayanan2017SimpleAS}. The uncertainty learned by standard network training may be particularly effective in terms of ranking, and can serve as a strong baseline for detecting misclassified and OOD examples~\citep{Hendrycks2017ABF}. Several simple regularization methods commonly used for improving performance have been shown to effectively improve the model uncertainty as well. These include early stopping or more advanced instance-wise early stopping~\citep{Geifman2019BiasReducedUE}, label smoothing~\citep{Mller2019WhenDL}, focal loss~\citep{Mukhoti2020CalibratingDN}, dropout~\citep{Srivastava2014DropoutAS} and data augmentation methods such as mixup~\citep{Thulasidasan2019OnMT} and Augmix~\citep{Hendrycks2020AugMixAS}. 
$$
\lambda \sim Beta(\alpha, \alpha)
$$

More advanced methods for improving performance are also demonstrated to improve uncertainty. Adversarial training~\citep{Goodfellow2015ExplainingAH, Kurakin2017AdversarialML, Madry2018TowardsDL}, originally designed to improve the robustness of deep neural networks against adversarial examples~\citep{Goodfellow2015ExplainingAH}, have been shown to improve the uncertainty~\citep{Lakshminarayanan2017SimpleAS}. Knowledge distillation can also be viewed as a regularization method with the aid of auxiliary models and has been shown to improve uncertainty in a diverse form, such as self-distillation~\citep{Kim2021SelfKnowledgeDW}, dropout distillation~\citep{Bul2016DropoutD, Gurau2018DropoutDF} and ensemble distillation~\citep{Mariet2020DistillingEI}. Recent years have seen a boom in pre-training, which trains the model on a large corpus without any labels, or in a so-called self-supervised scheme. Pre-training can benefit the learning of ``universal representations'' that transfers to multiple domains~\citep{Rebuffi2017LearningMV}, and can be viewed as a better initialization of a deep neural network. Therefore, it is not surprising that pre-training can improve the model uncertainty and robustness, even when the second-stage learning (or fine-tuning) is happening on a sufficiently large dataset where pre-training fails to improve the performance significantly compared to directly training on it~\citep{Hendrycks2019UsingPC}.

There are also a variety of regularization methods specifically designed for uncertainty learning, such as penalizing low-entropy predictive distribution in the training objective~\citep{Pereyra2017RegularizingNN}, interpolating the predictive distribution and the one-hot label using uncertainty score~\citep{Devries2018LearningCF}, incorporating OOD examples into training and enforcing the model to produce low-confidence predictions on them~\citep{Lee2018TrainingCC}, variance-weighted variation of label smoothing~\citep{Seo2019LearningFS}. \citet{Moon2020ConfidenceAwareLF} shows that penalizing the ranking difference between the uncertainty yielded by predictions at one training step and the moving-averaged predictions at multiple training steps can improve uncertainty learning, particularly effective for ordinal ranking. \citet{Maddox2019ASB} shows that the mean and variance of network weights across multiple training steps can serve as a good prior for their uncertainty distributions, thus building an efficient approximation for BNNs. One can also utilize an alternative network or alternative network modules to specialize in uncertainty learning while leaving the original network intact~\citep{Corbire2019AddressingFP, Geifman2019SelectiveNetAD}.

Note that these regularization methods are not necessarily orthogonal to each other, and may even hurt the uncertainty or accuracy when combined together. For example, it is well-known that label smoothing and knowledge distillation are not compatible with each other~\citep{Mller2019WhenDL}. And ensembling (see Section~\ref{sect:ensemble-uncertainty}) combined with data augmentation methods such as mixup can also hurt the model calibration~\citep{Wen2021CombiningEA}.

% Over-confidence. In-distribution (or mis-calibrated)

% OOD (Covariate shift?)

% Label shift
% * No longer valid to do calibration?
\subsection{Ensemble uncertainty}
\label{sect:ensemble-uncertainty}
In this section, we specifically focus on those methods that achieve better uncertainty quality by aggregating multiple classic uncertainty estimates. In essence, any randomness or perturbation in the input, sampling, weights, and optimization of deep neural networks during training or inference can be utilized to generate multiple uncertainty estimates for aggregation~\citep{Renda2019ComparingES}. We thus roughly partition the diverse ensembling methods into three families based on the origins of such randomness or perturbation.

\smallsection{Model ensemble}
Probably the most well-known uncertainty ensemble method is Monte-Carlo Dropout (MC-dropout)~\citep{Gal2016DropoutAA}. The idea is to utilize dropout to randomize the inference process of a network and average multiple stochastic predictions of one example to generate a better uncertainty estimate. Despite its simplicity, MC-dropout is shown to be akin to the Gaussian process and thus can be an efficient approximation of BNNs~\citep{Gal2016DropoutAA}, and has been widely used in practice. \citet{Bachman2014LearningWP} generalized such an idea using any network modules to randomize the inference process. 

Another well-known ensemble method is typically referred to as deep ensemble~\citep{Lakshminarayanan2017SimpleAS}. Here the randomness originates from the initialization of the deep neural networks, which is typically sampled from Gaussian distributions. One can thus train multiple deep neural networks on the same training set, and average their predictive distributions to obtain a better uncertainty estimate. It is possible to further promote the diversity of deep ensemble by combining the randomness in the data sampling process, e.g., bagging and boosting~\citep{Livieris2021OnET}. But in general cases, such a combination may not necessarily improve the uncertainty estimates and sometimes hurts the performance~\citep{Lee2015WhyMH, Lakshminarayanan2017SimpleAS}. Deep ensemble is widely used in practice, but may suffer from both increased training cost and inference cost. To reduce the inference cost, one can distill the knowledge of the ensemble into a single and small network~\citep{Hinton2015DistillingTK, Mariet2020DistillingEI, Nam2021DiversityMW}. To reduce the training cost, instead of training multiple networks, one can also only train multiple modules and share other modules. For example, \citet{Kim2018AttentionbasedEF} shows ensemble multiple attention modules trained with different attention masks can improve the uncertainty for image retrieval. \citet{Wen2020BatchEnsembleAA} generalizes such an idea by composing each weight matrix in a network based on the Hadamard product between a shared matrix among all ensemble members and a low-cost matrix unique to each member. % \citet{Wen2018FlipoutEP} use 

% \note{Add batch ensemble}

\smallsection{Input ensemble}
We have already mentioned the idea of using bagging or boosting to promote diversity in ensemble. This on its own can in fact be viewed as input ensemble, namely the randomness originates from either the sampling of the inputs, which can be utilized in training, or perturbation and data augmentation of the inputs, which can be utilized in both training and inference. For example, during training, \citet{Nanni2019DataAF} utilizes data augmentation to build an ensemble for bioimage classification while \citet{Guo2015DeepCE} utilizes data augmentation for deep ensemble in object detection. 

During inference, aggregating by data augmentation is even more widely used, particularly in medical image processing~\citep{Wang2018AutomaticBT, Wang2019AleatoricUE}. Such a technique is also commonly referred to as test-time augmentation~\citep{Ayhan2018TesttimeDA}. Empirical analyses found that test-time augmentation is quite sensitive to the data augmentation being used~\citep{Shanmugam2020WhenAW}. Methods are thus proposed to learn appropriate data augmentation, for example, that customizes for each test input individually~\citep{Kim2020LearningLF}.

\smallsection{Optimization ensemble}
The randomness of the optimization of a deep neural network can also be utilized to generate an ensemble. Snapshot ensemble~\citep{Huang2017SnapshotET} utilizes the fact that the weight landscape of a deep neural network is non-convex and the weights may traverse multiple local minima during optimization. They thus propose repeatedly decreasing and increasing the learning rate to let the optimization converge multiple times. Each converged network checkpoint can then serve as a member of the ensemble. \citet{Yang2020AutoEnsembleAA} adapts such an idea to adaptive learning rate schedulers. Fast Geometric Ensembling (FGE)~\citep{Garipov2018LossSM} and Stochastic Weight Averaging (SWA)~\citep{Izmailov2018AveragingWL} share a similar idea with snapshot ensemble, albeit the specific strategy to sample checkpoints along the training trajectory may differ. Hyperparameter ensemble~\citep{Wenzel2020HyperparameterEF} is a general method for optimization ensemble which uses AutoML to search for multiple hyperparameters for ensembling.

% Ensemble:
% * Dropout ensemble (mc-dropout)
% * Snapshot ensemble
% * Batch ensemble
% * Deep ensemble
% * test-time augmentation
% Refer to `COMBINING ENSEMBLES AND DATA AUGMENTATION CAN HARM YOUR CALIBRATION`: `By averaging predictions, ensembles can rule out individual mistakes (Lakshminarayanan et al., 2017; Ovadia et al., 2019). Additional work has gone into efficient ensembles such as MC-dropout (Gal and Ghahramani, 2016), BatchEnsemble, and its variants (Wen et al., 2018; 2020; Dusenberry et al., 2020; Wenzel et al., 2020).`

\subsection{``Deep'' uncertainty}
We now move from the classic definition of uncertainty, namely predictive distribution and its variants, to a more generalized definition of uncertainty, especially those that are unique to the training or inference process of deep neural networks. We roughly partition these uncertainty measures into two groups, namely those based on inference dynamics, and those based on training dynamics.

\smallsection{Inference dynamics}
During the inference of deep neural networks, typically only the outputs are desired. However, it is possible to mine other forms of measures associated with the network inference process as uncertainty. For example, \citet{Oberdiek2018ClassificationUO} and \citet{Lee2020GradientsAA} measure the uncertainty of a prediction using gradient information. When predicting on a single example, the gradients of the network weights with respect to the loss are calculated, where the label is simply the predicted class. The uncertainty can then be defined as the norm of the gradients.

When the training set or some training examples are available during inference, one can measure the uncertainty by quantifying the ``similarity'' between the test example and the training examples~\citep{Raghu2019DirectUP, Ramalho2020DensityEI}. In essence, such methods build a density estimation in the input space and thus can reject those test inputs that are likely to be off-distribution. \citet{Amersfoort2020SimpleAS} adapts this idea by maintaining only a set of representative training examples (or their feature vectors) called centroids. The uncertainty is then quantified as the distance between the test input and the centroid that is closest to it. Note that such uncertainty measures dispense with the need to predict the label. Alternatively, \citet{Jiang2018ToTO} defines the uncertainty as the ratio between the distance from the test input to the closest centroid and the distance from the test input to the centroid associated with the class predicted by the model, which can be more robust. These methods may be inherently connected to few-shot learning methods such as ProtoNet~\citep{Snell2017PrototypicalNF} where the training set is also directly used to help the inference.

% \note{Adversarial stability and augmentation stability (Based on input randomness) (my work)}

Finally, there exist some intriguing behaviors of the inference process of deep neural networks that can be leveraged to quantify the uncertainty. \citet{Baldock2021DeepLT} observed that through the forward propagation in a deep neural network with multiple layers, the prediction of some examples may already be determined after only a few layers. Here the intermediate predictions are made by k-nearest neighbors classifiers on the hidden representations. They thus define an uncertainty measure called Prediction Depth (PD). It is shown that the prediction depth may be closely correlated with the margin of the final predictive distribution and examples with large prediction depth may be more difficult. Second, it is known that the inference of deep neural network is vulnerable to adversarial perturbation. It is observed that the predictions on some examples are more resistant to adversarial perturbation. Therefore, one can define the smallest perturbation size required to change the model's prediction as an uncertainty measure~\citep{Carlini2019DistributionDT}. Such a metric, typically referred as minimum adversarial perturbation~\citep{Carlini2017TowardsET} or adversarial input margin~\citep{Baldock2021DeepLT}, may also be closely correlated with the uncertainty defined on the predictive distribution.

\smallsection{Training dynamics}
Deep neural networks may exhibit even more intriguing behaviors during training. It is well-known that deep neural networks can perfectly fit even pure noise~\citep{Zhang2017UnderstandingDL}, which raises the question of whether deep networks simply ``memorize'' the training set even on real datasets. However, through a careful investigation of the predictions on individual training examples, \citet{Arpit2017ACL} finds that data examples are not learned at the same pace during training. Those real data examples are learned first, while those random data, either with random input or random labels, are learned late. Further, they also found that simple data examples are learned earlier than those difficult data examples. These observations demonstrate that deep neural networks are not simply memorizing data since they appear to be aware of the content and semantics. \citet{Hacohen2020LetsAT} further shows such a learning order of training examples is consistent across different random initializations of a network and different model architectures. Even more intriguingly, they find that such a consistent learning order is not observed for non-parametric classifiers such as AdaBoost. They also observed that when trained on synthetic datasets where the images are different rotation or colorization of Gabor patches, such a consistent learning order disappears as well. They thus hypothesize that such an intriguing behavior may originate from the interplay between deep neural networks and natural datasets with recognizable patterns and semantics. 

\citet{Toneva2019AnES} observed that certain training examples are frequently forgotten during training, which means that they can be first predicted correctly, then incorrectly. The frequency of such forgotten events is shared across different neural architectures. When removing those least forgettable examples from training, model performance can be largely maintained  Nevertheless, \citet{Dong2021DataQM} shows an opposite trend exists in adversarial training, where those most forgettable examples~\footnote{Here a correct prediction is defined under adversarial perturbation.} may be removed without degrading performance, and sometimes even improving it. The variance of the gradient of a data example during training is also shown to be strongly correlated with its difficulty~\citep{Agarwal2022EstimatingED}.

% Training dynamics
% * Learning order
% * Learning stability
% * Variance of input graidents `Estimating Example Difficulty using Variance of Gradients`

% Lipschitzness

%% file: sections/4-usage.tex
\section{Utilize Uncertainty for Better Performance}
\label{sect:usage}
Starting from this section, we demonstrate that the uncertainty of deep neural networks can be utilized to improve the robustness and efficiency of learning systems in a variety of realistic applications.

\subsection{Uncertainty for robust learning}
We first focus on the application of uncertainty in robust learning. Here we will skip the detection of OOD examples as it is often a standard application of model uncertainty. Instead, we will discuss how to utilize uncertainty in learning with noisy labels and adversarially robust learning.

\smallsection{Learning with noisy labels}
One straightforward idea for learning with noisy labels is to identify those labels in the training data that are likely to be incorrect. This naturally calls the necessity of uncertainty estimates, which is often referred to as Confidence Learning (CL)~\citep{Northcutt2021ConfidentLE} in noisy learning regime. Classic uncertainty can, of course, be utilized to identify noisy labels, but may be inferior since sufficiently trained deep neural networks can memorize the labels and thus be over-confident. To combat this, one can simply perform early stopping to obtain a better uncertainty~\citep{Liu2020EarlyLearningRP}. This implicitly leverages the training dynamics of deep neural networks, namely noisy data will be learned late during training. By further combining the observation of prediction depth, namely noisy data will be learned in later layers of a deep neural network, one can early stop different parts of the network at different training checkpoints. In specific, early layers will be trained first and later layers will be progressively trained with few epochs, while the early layers will be frozen~\citep{Bai2021UnderstandingAI}. \citet{Xia2021RobustEH} further exploits this idea by dividing all network weights into those that are important for fitting clean labels and those that are important for fitting noisy labels based on gradient norm. The former weights are updated regularly while the latter are simply penalized with weight decay. This can be viewed as early stopping the training of the network in a parameter-wise manner. \citet{Han2020SIGUAFM} also considers suppressing the learning on data examples that are likely to have noisy labels during training to avoid memorizing noisy labels.

Instead of seeking better output uncertainty estimates, several methods directly utilize the uncertainty yielded by training dynamics of deep neural networks as a metric to identify noisy labels. For example, \citet{TCPCurriculum} proposes using Time-Consistency Prediction (TCP) to select clean data, which is simply the stability of the prediction on an example throughout training. Slightly differently, \citet{Pleiss2020IdentifyingMD} proposes to use the averaged margin of an example through training to select clean data.

On top of detecting noisy labels, uncertainty estimates are also important in more advanced and sophisticated noisy learning methods. For example, Co-teaching~\citep{Han2018CoteachingRT} trains two networks simultaneously where one network is trained on the potentially clean labels selected by its peer network in a mini-batch. A reasonably good uncertainty estimate for noisy labels is necessary here. DivideMix~\citep{Li2020DivideMixLW} further develops this idea by using one network to divide the training set into a clean partition and a noisy partition. The peer network is trained on the clean partition, along with the noisy partition without labels in a semi-supervised manner. DivideMix has achieved state-of-the-art on multiple noisy learning benchmarks.
% Input robustness

% \smallsection{Spurious correlation}
% Output robustness

% \subsection{OOD detection}
% Input robustness

\smallsection{Adversarially robust learning}
Adversarial training~\citep{Goodfellow2015ExplainingAH, Madry2018TowardsDL} is so far the most effective way to enhance the robustness of deep neural networks against adversarial examples. However, the robust accuracy achieved on small benchmark datasets such as CIFAR-10~\citep{Krizhevsky2009LearningML} or CIFAR-100~\citep{Krizhevsky2009LearningML} is still unsatisfactory, not to mention larger datasets such as ImageNet~\citep{Russakovsky2014ImageNetLS}, especially against those strong adversarial attacks such as AutoAttack~\citep{Croce2020ReliableEO}. Surprisingly, recent studies found that robustness achieved by adversarial training can be greatly boosted if the deep neural networks are trained on additional data, which is either selected from large unlabeled data corpus~\citep{Uesato2019AreLR, Carmon2019UnlabeledDI} or simply generated by generative models~\citep{Sehwag2021RobustLM, Gowal2021ImprovingRU}, thus requiring minimal human effort. Because those additional data examples may be far away from the original data distribution, selecting high-quality additional data is demonstrated to be crucial in this process~\citep{Uesato2019AreLR, Gowal2020UncoveringTL}. The typical metric used to select additional data is the classic uncertainty score yielded by a classifier trained on the in-distribution clean data. \citet{Dong2021DataQM} suggests that it may be better to use training stability, namely the frequency of predictions on an example being correct throughout training. However, how to build an uncertainty metric to select high-quality additional data remains an open and important problem.

% * Extra data including unlabeled data and generated data helps adversarially robust deep learning, although generate data doesn't help standard learning (work seen today explained why: `why do artificially generated data help adversarial robustness` Not public yet.)
% Select high-quality extra data requires to learn uncertainty.
% * `Recent works have shown that the robustness achieved by adversarial training can be significantly boosted by introducing additional data that are either unlabeled (Uesato et al., 2019; Carmon et al., 2019) or generated by generative models (Sehwag et al., 2021; Gowal et al., 2021). Selecting high-quality data is crucial in this process as the additional data may be far from the original data distribution (Uesato et al., 2019; Gowal et al., 81 2020). `

Some intriguing problems in adversarially robust learning may also require learning good uncertainty estimates. For example, robust overfitting~\citep{Rice2020OverfittingIA} is a well-known problem of adversarial training, which refers to the phenomenon that during adversarial training, the robust accuracy on the test set will unexpectedly start to decrease after a certain number of training steps. Such overfitting occurs consistently across different datasets, training settings, adversary settings, and neural architectures. However, when conducting standard training on the same dataset, such an overfitting phenomenon is not observed. Recent work shows that robust overfitting may originate from the label noise implicitly exists in adversarial training, which is induced by the mismatch between the true conditional distribution and the label distribution of the adversarial examples used for training~\citep{Dong2021LabelNI}. Therefore, a straightforward solution to robust overfitting is to recover the true conditional distribution using uncertainty estimates generated by a model, and is demonstrated to significantly alleviate robust overfitting~\citep{Chen2021RobustOM}.

% --------------------------------------------------------
\subsection{Uncertainty for efficient learning}
In this section, we show that the uncertainty of deep neural networks can be utilized to improve efficiency of deep neural networks. Here we focus on two types of efficiency concerns: data efficiency, namely to reduce the human effort on data annotation in training deep neural networks, and model efficiency, namely to reduce the computation cost of the training and inference of deep neural networks.

\subsubsection{Data efficiency}
\smallsection{Semi-supervised learning}
Semi-supervised learning tackles the challenge in real-world applications where only a limited number of labeled data is available, while the vast majority of data is unlabeled. A natural solution to semi-supervised learning is incorporating the labels predicted by a classifier on unlabeled data into training, which are typically referred to as pseudo-labels~\citep{Lee2013PseudoLabelT}. Such a process can be repeatedly conducted when a better classifier is trained after more pseudo-labeled data is available, which is known as bootstrapping or self-training~\citep{McClosky2006EffectiveSF}. 

However, since the pseudo-labels predicted by a classifier are very likely to be incorrect, it is crucial to define a reliable uncertainty measure to select those correct pseudo-labels. Classic uncertainty can be utilized here but may inevitably suffer from over-confidence~\citep{Guo2017OnCO} and memorization~\cite{Zhang2017UnderstandingDL}. The recent development of self-training has seen an increasing trend of ``uncertainty-aware'' methods. For example, instead of using classic uncertainty, \citet{Mukherjee2020UncertaintyawareSF} uses MC-dropout to select pseudo-labels for self-training. They also design a loss function that incorporates the uncertainty on the correctness of pseudo-labels into training, where those selected pseudo-labels that are more likely to be correct will be more focused on. \citet{Rizve2021InDO} further shows that self-training using MC-dropout as an uncertainty measure along with other careful designs can compete with much more sophisticated semi-supervised learning methods such as consistency regularization~\citep{Laine2016TemporalEF, Tarvainen2017MeanTA} and mixed methods~\citep{Berthelot2019MixMatchAH}, albeit enjoying the advantage that no data augmentation is explicitly required. \citet{Zhou2020TimeConsistentSF} uses the consistency of the model prediction through training as an uncertainty measure in self-training. The hypothesis here is those model predictions that are largely invariant during training are more likely to be correct. This again is reminiscent of the intriguing training dynamics of deep neural networks as mentioned before.

% Entropy-regularized semi-supervised learning?

\smallsection{Weakly-supervised learning}
Weakly-supervised learning is even more challenging than semi-supervised learning in that the limited number of labels are not absolutely correct. In most cases, they are generated by a set of pre-defined rules or provided by annotators without domain expertise, which are referred to as weak supervision. Under such circumstances, a reliable uncertainty estimate becomes more important in not only selecting pseudo-labels predicted by classifiers on the unlabeled set, but also selecting those labels given by weak supervision that are likely to be correct. \citet{Mekala2022LOPSLO} provides a comprehensive analysis of uncertainty estimates in weakly-supervised text classification. They found that among multiple uncertainty measures including classic uncertainty, prediction stability and MC-dropout, the learning order performs the best for selecting pseudo-labels given by weak supervision. As mentioned before, the learning order here refers to the consistent order of learning real and noisy data by deep neural networks during training. Because incorrect pseudo-labels are learned late by the model, the epoch at which the model prediction matches the pseudo-label can be an effective metric to distinguish correct and incorrect labels.

% \smallsection{Active learning}
% Data efficiency

\subsubsection{Model efficiency}
\smallsection{Model compression and Knowledge distillation} To excel in realistic tasks, deep neural networks often have to be excessively large in capacity. This brings significant computation burden in both training deep neural networks and using deep neural networks for predictions. How to compress a large deep neural network while maintaining its performance is thus gaining increasing attention. One well-known method to compress deep neural networks is knowledge distillation~\citep{Hinton2015DistillingTK}, namely a small network called student is trained using the predictions provided by a large network called teacher, rather than using the original labels in the training set. Despite its success, why teacher predictions can help student learning has always been a mystery. A recent finding, that the student predictions on the training or test set can often disagree with the teacher predictions on a large number of examples~\citep{Stanton2021DoesKD}, further mystifies knowledge distillation. 

Toward better understanding and improving knowledge distillation, a significant effort in theoretical analyses has been made in recent works. In specific, \citet{Menon2021ASP} shows that the predictive distribution provided by the teacher can improve the generalization of students because the predictive distribution is a better approximation to the true conditional distribution than one-hot labels, where true conditional distribution as supervision can reduce the variance. \citet{Dao2021KnowledgeDA} shows that the distance between teacher predictions and the true conditional distribution can directly bound the student accuracy. Based on this understanding, to improve knowledge distillation we essentially require the teacher to learn better uncertainty estimates in terms of its recovery of the true conditional distribution. Recent work has thus proposed to directly optimize the teacher to learn the true conditional distribution, which is dubbed as student-oriented teacher training and has achieved better student performance~\cite{Dong2022SoTeacherAS}.

\smallsection{Adaptive inference time}
Deep neural networks often consist of a great many layers. The inference process can thus be computationally prohibitive as the calculation of each layer has to be made sequentially, one after another. An idea to reduce the inference cost is to early stop the inference process of some examples when the outputs of the intermediate layers are already informative enough to make predictions, which is known as adaptive inference time in sequence processing and classification~\citep{Graves2016AdaptiveCT}. Such a strategy leverages the inference dynamics of deep neural networks where easy examples may have a small prediction depth. 

To ensure an accurate early stopping, a reliable uncertainty estimate is important here to determine when the early predictions are likely to be correct. The entropy of the predictive distribution at early classification heads is widely used in adaptive inference methods across applications in computer vision and natural language processing, albeit the specific designs of the network architecture or training strategies differ.  
BranchyNet~\citep{Teerapittayanon2016BranchyNetFI} inserts multiple early-exit classification heads between intermediate layers and trains the network to minimize the weighted sum of the loss functions defined at all classification heads.  
Multi-Scale DenseNet (MSDNet)~\citep{Huang2017MultiScaleDN} refines this idea by altering the network architecture such that feature representations at multiple scales can be utilized jointly to determine an early exit. In sequence classification where transformers dominate, adaptive inference time can be built into the network architecture~\citep{Dehghani2018UniversalT, Xin2020DeeBERTDE}. FastBert~\citep{Liu2020FastBERTAS} improves the early-exit accuracy by introducing self-distillation on intermediate classification heads. \citet{Schwartz2020TheRT} improves the uncertainty estimate for early exits by using temperature scaling to calibrate the predictive distribution and achieves a better speed-accuracy trade-off.

% \smallsection{Curriculum learning}
% Training efficiency

%% file: sections/6-conclusion.tex
\section{Conclusion}
\label{sect:conclusion}
In this survey, we review a wide spectrum of uncertainty measures we can define for deep neural networks. These include the classic definitions of uncertainty such as those based on the predictive distribution, and those definitions of uncertainty that are closely connected to the training and inference dynamics of deep neural networks. We show that these uncertainty measures can be leveraged in realistic applications across computer vision and natural language processing, to improve the robustness and efficiency of learning systems. We believe there are more scenarios where reliable uncertainty estimates are crucial to the performance. We also believe the increasingly popular use cases of uncertainty estimates beyond interpretability and transparency will in turn facilitate more opportunities in uncertainty learning of deep neural networks.